%% file: main.tex
\documentclass{article}

\usepackage{microtype}
\usepackage{graphicx}
\usepackage{subfigure}
\usepackage{booktabs}
\usepackage{enumitem}
\usepackage{hyperref}
\usepackage{url}
\usepackage{graphicx}
\usepackage{multicol}
\usepackage{multirow}
\usepackage{xcolor}
\usepackage{amssymb}
\usepackage{titling}
\usepackage{caption}
\usepackage{ulem}

\usepackage[accepted]{icml2022}

\input{math_commands.tex}

\icmltitlerunning{Domain Adaptation for Time Series Forecasting via Attention Sharing}

\begin{document}
\twocolumn[
\icmltitle{Domain Adaptation for Time Series Forecasting via Attention Sharing}



\icmlsetsymbol{equal}{*}

\begin{icmlauthorlist}
\icmlauthor{Xiaoyong Jin}{amz}
\icmlauthor{Youngsuk Park}{amz}
\icmlauthor{Danielle C. Maddix}{amz}
\icmlauthor{Hao Wang}{amz,rutgers}
\icmlauthor{Yuyang Wang}{amz}
\end{icmlauthorlist}

\icmlaffiliation{amz}{AWS AI Labs}
\icmlaffiliation{rutgers}{Rutgers University}

\icmlcorrespondingauthor{Xiaoyong Jin}{jxiaoyon@amazon.com}

\icmlkeywords{Machine Learning, ICML}

\vskip 0.3in
]



\printAffiliationsAndNotice{} 

\begin{abstract}


Recently, deep neural networks have gained increasing popularity in the field of time series forecasting. A primary reason for their success is their ability to effectively capture complex temporal dynamics across multiple related time series. The advantages of these deep forecasters only start to emerge in the presence of a sufficient amount of data. This poses a challenge for typical forecasting problems in practice, where there is a limited number of time series or observations per time series, or both. To cope with this data scarcity issue, we propose a novel domain adaptation framework, Domain Adaptation Forecaster (DAF).  DAF leverages statistical strengths from a relevant domain with abundant data samples (source) to improve the performance on the domain of interest with limited data (target). In particular, we use an attention-based shared module with a domain discriminator across domains and private modules for individual domains. We induce domain-invariant latent features (queries and keys) and retrain domain-specific features (values) simultaneously to enable joint training of forecasters on source and target domains. A main insight is that our design of aligning keys allows the target domain to leverage source time series even with different characteristics.  Extensive experiments on various domains demonstrate that our proposed method outperforms state-of-the-art baselines on synthetic and real-world datasets, and ablation studies verify the effectiveness of our design choices.

\end{abstract}


\section{Introduction}

Similar to other fields with predictive tasks, time series forecasting has recently benefited from the development of deep neural networks \citep{flunkert_deepar_2020, borovykh_conditional_2017, oreshkin_n-beats_2020}, ultimately toward final decision making systems like retail~\citep{bose2017probabilistic}, resource planning for cloud computing~\citep{park2019linear}, and optimal control vehicles~\citep{kim2020optimal, park_learning_2022}. In particular, based on the success of Transformer models in natural language processing \citep{vaswani_attention_2017}, attention models have also been effectively applied to forecasting \citep{li_enhancing_2019, lim_temporal_2019, zhou_informer_2021, xu_autoformer_2021}.   
While these deep forecasting models excel at capturing complex temporal dynamics from a sufficiently large time series dataset, it is often challenging in practice to collect enough data. 

A common solution to the data scarcity problem is to introduce another dataset with abundant data samples from a so-called source domain related to the dataset of interest, referred to as the target domain. For example, traffic data from an area with an abundant number of sensors (source domain) can be used to train a model to forecast the traffic flow in an area with insufficient monitoring recordings (target domain).  However, deep neural networks trained on one domain can be poor at generalizing to another domain due to the issue of domain shift, that is, the distributional discrepancy between domains \citep{wang_bridging_2021}.

Domain adaptation (DA) methods attempt to mitigate the harmful effect of domain shift by aligning features extracted across source and target domains \citep{ganin_domain-adversarial_2016, bousmalis_domain_2016, hoffman_cycada_2018, bartunov_few-shot_2018,CIDA}. Existing approaches mainly focus on classification tasks, where a classifier learns a mapping from a learned domain-invariant latent space to a fixed label space using source data. Consequently, the classifier depends only on common features across domains, and can be applied to the target domain \citep{wilson_survey_2020}. 

\begin{figure*}[h]
    \centering
    \includegraphics[scale=0.4]{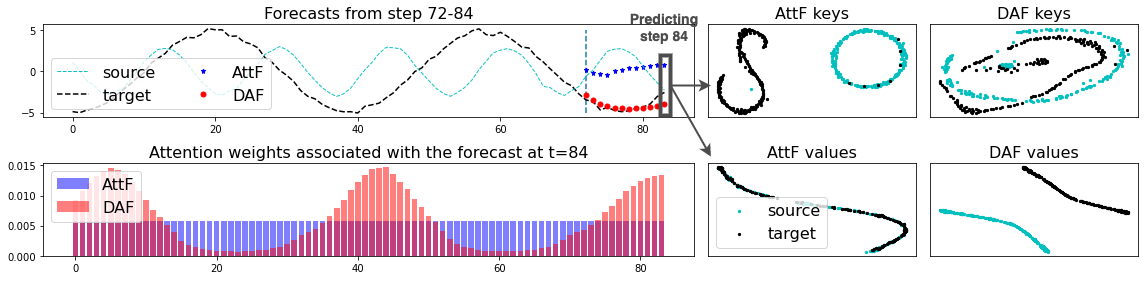}
    \caption{Forecasts of single-domain attention-based forecaster (AttF) and our cross-domain forecaster (DAF) on synthetic data. Sample forecasts from steps 72-84 on the target domain where our DAF is also trained on the source domain \textit{(top left}). Bar plot of the weights on the context history of the attention distributions of AttF and DAF associated with forecasting step 84 (\textit{bottom left}). Attention keys (\textit{top right}) and values (\textit{bottom right}) of AttF and DAF after dimension reduction to 2D. The keys and values of AttF in the source domain are generated by simply applying AttF model trained on target data to the source data. The strategy of aligning keys rather than values between source and target domains in our DAF captures the correct attention weights, as illustrated by the accurate forecasts compared to AttF (cf. red dots vs. blue dots from steps 72-84). 
    } 
    
    \label{fig:visual}
\end{figure*}

There are two main challenges in directly applying existing DA methods to time series forecasting.
First, due to the temporal nature of time series, evolving patterns within time series are not likely to be captured by a representation of the entire history.   
Future predictions may depend on local patterns within different time periods, and a sequence of \textit{local representations} can be more appropriate than using the \textit{entire history} as done with most conventional approaches.  
Second, the \textit{output space} in forecasting tasks is not fixed across domains in general since a forecaster generates a time series following the input, which is domain-dependent, e.g. kW in electrical source data vs. unit count in stock target data.  Both domain-invariant and domain-specific features need to be extracted and incorporated in forecasting to model domain-dependent properties so that the data distribution of the respective domain is properly approximated.
Hence, we need to carefully design the type of features to be shared or non-shared over different domains, and to choose a suitable architecture for our time-series forecasting model.

    We propose to resolve the two challenges using an attention-based model \cite{vaswani_attention_2017} equipped with domain adaptation. First, for evolving patterns, attention models can make dynamic forecasts based on a combination of values weighted by time-dependent query-key alignments. Second, as the alignments in an attention module are independent of specific patterns, the queries and keys can be induced to be domain-invariant while the values can stay domain-specific for the model to make domain-dependent forecasts. \Figref{fig:visual} presents an illustrative example of a comparison between a conventional attention-based forecaster (AttF) and its counterpart combined with our domain adaptation strategy (DAF) on synthetic datasets with sinusoidal signals. While AttF is trained using limited target data, DAF is jointly trained on both domains. By aligning the keys across domains as the top rightmost panel shows, the context matching learned in the source domain helps DAF generate more reasonable attention weights that focus on the same phases in previous periods of target data than the uniform weights generated by AttF in the bottom left panel. The bottom right panels illustrate that the single-domain AttF produces the same values for both domains as the input is highly overlapped, while DAF is able to generate distinct values for each domain.   As a result, the top left panel shows that DAF produces more accurate domain-specific forecasts than AttF does. 

In this paper, we propose the Domain Adaptation Forecaster (DAF), a novel method that effectively solves the data scarcity issue in the specific task of time series forecasting by applying domain adaptation techniques via attention sharing.  
The main contributions of this paper are:
\begin{enumerate}[noitemsep, topsep=0pt, leftmargin=1.5em]
    \item In DAF, we propose a new architecture that properly induces and combines domain-invariant and domain-specific features to make multi-horizon forecasts for source and target domains through a shared attention module. To the best of our knowledge, our work provides the first end-to-end DA solution specific for multi-horizon forecasting tasks with adversarial training.
    \item{We demonstrate that DAF outperforms state-of-the-art single-domain forecasting and domain adaptation baselines in terms of accuracy in a data-scarce target domain through extensive synthetic and real-world experiments that solve cold-start and few-shot forecasting problems.}
    \item{We perform extensive ablation studies to show the importance of the domain-invariant features induced by a discriminator and the 
    retrained domain-specific features in our DAF model, 
    and that our designed sharing strategies with the discriminator result in better performance than other potential variants.} 
\end{enumerate}

\section{Related Work}\label{sec:liter}
Deep neural networks have been introduced to time series forecasting with considerable successes \citep{flunkert_deepar_2020, borovykh_conditional_2017, oreshkin_n-beats_2020, wen_multi-horizon_2017, wang_deep_2019, sen_think_2019, rangapuram_deep_2018, park_learning_2022, kan2022multivariate}. In particular, attention-based transformer-like models \citep{vaswani_attention_2017} have achieved state-of-the-art performance \citep{li_enhancing_2019, lim_temporal_2019, wu_adversarial_2020, zhou_informer_2021}. A downside to these sophisticated models is their reliance on a large dataset with homogeneous time series to train. Once trained, the deep learning models may not generalize well to a new domain of exogenous data due to domain shift issues \citep{wang_issues_2005, purushotham_variational_2017, wang_bridging_2021}. There are several robust forecasting methods \citep{yoon2022robust} in terms of defending adversarial attacks, but does not explicitly cover general domain shifts beyond adversarial ones.

To solve the domain shift issue, domain adaptation has been proposed to transfer knowledge captured from a source domain with sufficient data to the target domain with unlabeled or insufficiently labeled data for various tasks \citep{motiian_few-shot_2017, wilson_survey_2020, ramponi_neural_2020}. In particular, sequence modeling tasks in natural language processing mainly adopt a paradigm where large transformers are successively pre-trained on a general domain and fine-tuned on the task domain \citep{devlin_bert_2019,  han_unsupervised_2019, gururangan_dont_2020, rietzler_adapt_2020, yao_unsupervised_2020}. 
It is not immediate to directly apply these methods to forecasting scenarios due to several challenges. First, it is difficult to find a common source dataset in time series forecasting to pre-train a large forecasting model. Second, it is expensive to pre-train a different model for each target domain. Third, the predicted values are not subject to a fixed vocabulary, heavily relying on extrapolation. Lastly, there are many domain-specific confounding factors that cannot be encoded by a pre-trained model.

An alternative approach to pre-training and fine-tuning for domain adaptation is to extract domain-invariant representations from raw data \citep{ben-david_theory_2010, kivinen_domain_2011}. Then a recognition model that learns to predict labels using the source data can be applied to the target data. 
In their seminal works, \citet{ganin_unsupervised_2015, ganin_domain-adversarial_2016} propose DANN to obtain domain invariance by confusing a domain discriminator that is trained to distinguish representations from different domains. A series of works follow this adversarial training paradigm \citep{tzeng_adversarial_2017, zhao_adversarial_2018, alam_domain_2018, wright_transformer_2020,CIDA,GRDA}, and outperform conventional metric-based approaches \citep{long_learning_2015, chen_closer_2020, guo_multi-source_2020} in various applications of domain adaptation. However, these works do not consider the task of time series forecasting, and address the challenges in the introduction accordingly.

In light of successes in related fields, domain adaptation techniques have been introduced to time series tasks \citep{purushotham_variational_2017, wilson_multi-source_2020}.  \citet{cai_time_2021} aim to solve domain shift issues in classification and regression tasks by minimizing the discrepancy of the associative structure of time series variables between domains.  A limitation of this metric-based approach is that it cannot handle the multi-horizon forecasting task since the label is associated with the input rather than being pre-defined. \citet{hu_datsing_2020} propose DATSING to adopt adversarial training to fine-tune a pre-trained forecasting model by augmenting the target dataset with selected source data based on pre-defined metrics.  This approach lacks the efficiency of end-to-end solutions due to its two-stage nature. In addition, it does not consider domain-specific features to make domain-dependent forecasts. Lastly, \citet{ghifary_deep_2016, bousmalis_domain_2016, shi_genre_2018} make use of domain-invariant and domain-specific representations in adaptation. However, since these methods do not accommodate the sequential nature of time series, they cannot be directly applied to forecasting.

\section{Domain Adaptation in Forecasting}\label{sec:pre}
\paragraph{Time Series Forecasting}
Suppose a set of $N$ time series, and each consists of observations $z_{i,t} \in \mathbb{R}$, associated with optional input covariates $\xi_{i,t} \in \mathbb{R}^d$ such as price and promotion, at time $t$. In time series forecasting, given $T$ past observations and all future input covariates, we wish to make $\tau$ multi-horizon future predictions at time $T$ via model $F$:
\vspace{-.2cm}
\begin{equation}
\label{eqn:forecast}
z_{i,T+1}, \ldots, z_{i,T+\tau} = F(z_{i,1}, \ldots, z_{i,T}; \xi_{i,1}, \ldots, \xi_{i,T + \tau}).
\end{equation}
In this paper, we focus on the scenario where little data is available for the problem of interest while sufficient data from other sources is provided. For example, one or both of the number of time series $N$ and the length $T$ is limited. For notation simplicity, we drop the covariates  $\{\xi_{i,t}\}_{t=1}^{T + \tau}$ in the following. We denote the dataset $\gD=\{(\rmX_i,\rmY_i)\}_{i=1}^N$ with past observations $\rmX_i = [z_{i,t}]_{t=1}^{T}$ and future ground truths $\rmY_i = [z_{i,t}]_{t=T+1}^{T+\tau}$ for the $i$-th time series. We also omit the index $i$ when the context is clear.

\vspace{-.1cm}
\paragraph{Adversarial Domain Adaptation in Forecasting}
To find a suitable forecasting model $F$ in \eqref{eqn:forecast} on a data-scarce time series dataset, we cast the problem in terms of a domain adaptation problem, given that another ``relevant'' dataset is accessible. In the domain adaption setting, we have two types of data: source data $\mathcal{D}_{\gS}$ with abundant samples and target data $\mathcal{D}_{\gT}$ with limited samples. Our goal is to produce an accurate forecast on the target domain $\gT$, where little data is available, by leveraging the data in the source domain $\gS$. Since our goal is to provide a forecast in the target domain, in the remainder of the text, we use $T$ and $\tau$ to denote the target historical length and target prediction length, respectively, and also use the subscript $\gS$ for the corresponding quantities in the source data $\gD_{\gS}$, and likewise for $\gT$.

To compute the desired target prediction $\hat{\rmY}_i = [\hat{z}_{i,t}]_{t=T+1}^{T+\tau}$, $i=1, \ldots, N$, we optimize the training error on both domains jointly and in an adversarial manner in the following minimax problem:
\begin{equation}\label{eq:minimax_obj}
\begin{aligned}
    \min_{G_{\gS}, G_{\gT}} \max_{D} 
    \quad &\gL_{seq}(\gD_{\gS};G_{\gS}) 
    + \gL_{seq}(\gD_{\gT};G_{\gT})\\
    - &\lambda \gL_{dom}\left(\gD_{\gS},\gD_{\gT}; D,G_{\gS},G_{\gT}\right),
\end{aligned}
\end{equation}
where the parameter $\lambda \geq 0$ balances between the estimation error $\gL_{seq}$ and the domain classification error $\gL_{dom}$.
Here, $G_{\gS}, G_{\gT}$ denote sequence generators that estimate sequences in each domain, respectively, and $D$ denotes a discriminator that classifies the domain between source and target.

We first define the estimation error $\gL_{seq}$ induced by a sequence generator $G$ as follows: 
\begin{equation}\label{eq:l_seq}
\begin{aligned}
    \gL_{seq}(\gD; G) &= \\
    \sum_{i=1}^N
   \bigg(\frac{1}{T}&\sum_{t=1}^{T}l(z_{i,t}, \hat{z}_{i,t})
    +
   \frac{1}{\tau}\sum_{t=T+1}^{T+\tau}l(z_{i,t}, \hat{z}_{i,t})\bigg),
\end{aligned}
\end{equation}
where $l$ is a loss function and estimation $\hat{z}_{i,t}$ is the output of a generator $G$, and each term in \eqref{eq:l_seq} represents the error of input reconstruction and future prediction, respectively. 
Next, let $\gH =\{h_{i,t}\}_{i=1, t=1}^{N, T+\tau}$ be a set of some latent feature $h_{i,t}$ induced by generator $G$. Then, the domain classification error $\gL_{dom}$ in \eqref{eq:minimax_obj} denotes the cross-entropy loss in the latent spaces as follows:
\begin{equation}\label{eq:cross_entropy}
\begin{aligned}
    & \gL_{dom}(\gD_{\gS},\gD_{\gT}; D,G_{\gS},G_{\gT}) = \\
    & -\frac{1}{\vert\gH_{\gS}\vert}\sum_{h_{i,t} \in \gH_{\gS}}\log D(h_{i,t})\\
    & -\frac{1}{\vert\gH_{\gT}\vert}\sum_{h_{i,t} \in \gH_{\gT}}\log\left[1-D(h_{i,t})\right],
\end{aligned}
\end{equation}
where $\gH_{\gS} $ and $\gH_{\gT}$ are latent feature sets associated with the source $\gD_{\gS}$ and target $\gD_{\gT}$, and $|\gH|$ denotes the cardinality of a set $\gH$. 
The minimax objective \eqref{eq:minimax_obj} is optimized via adversarial training alternately. 
In the following subsections, we propose specific design choices for $G_{\gS}, G_{\gT}$ (see sub\secref{sec:sg}) and the latent features $ \mathcal{H}_{\gS}, \mathcal{H}_{\gT}$ (see sub\secref{sec:adv}) in our DAF model. 

\begin{figure*}[t]
    \centering
    \includegraphics[scale=0.4]{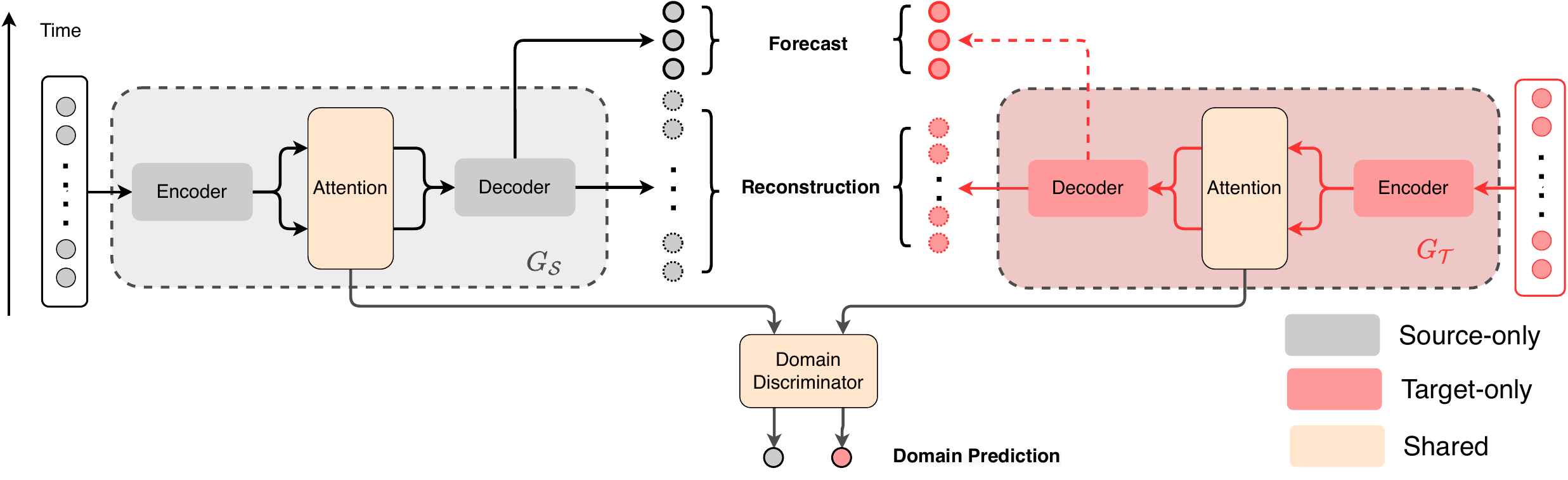}
    \caption{An architectural overview of DAF. The grey modules belong to the source domain, and red modules belong to target domain. The attention modules and domain discriminators shown in beige are shared by both domains. The model takes the historical portion of a time series as input, and produces a reconstruction of input and a forecast of the future time steps. 
    The domain discriminator is a binary classifier, and predicts the origin of an intermediate representation within the attention module, either the source or the target.}
    \label{fig:architecture}
\end{figure*}

\section{The Domain Adaptation Forecaster (DAF)}
\label{sec:method}

We propose a novel strategy based on attention mechanism to perform domain adaptation in forecasting. The proposed solution, the Domain Adaptation Forecaster (DAF), employs a sequence generator to process time series from each domain. Each sequence generator consists of an encoder, an attention module and a decoder.
As each domain provides data with distinct patterns from different spaces, we keep the encoders and decoders privately owned by the respective domain. The core attention module is shared by both domains for adaptation.
In addition to computing future predictions, the generator also reconstructs the input to further guarantee the effectiveness of the learned representations. \Figref{fig:architecture} illustrates an overview of the proposed architecture.

\subsection{Sequence Generators}\label{sec:sg}
In this subsection, we discuss our design of the sequence generators $G_{\gS}, G_{\gT}$ in \eqref{eq:minimax_obj}.
Since the generators for both domains have the same architecture, we omit the domain index of all quantities and denote either generator by $G$ in the following paragraphs by default. The generator $G$ in each domain processes an input time series $\rmX=[z_{t}]_{t=1}^T$ and generates the reconstructed sequence $\hat{\rmX}$ and the predicted future $\hat{\rmY}$.

\paragraph{Private Encoders} The private encoder transforms the raw input $\rmX$ into the pattern embedding $\rmP=[\rvp_t]_{t=1}^T$ and value embedding $\rmV=[\rvv_t]_{t=1}^T$. 
We apply a position-wise MLP with parameter $\bm{\theta}_v$ to encode input $\rmX=[z_{t}]_{t=1}^T$ for the value embedding $\rvv_t = \text{MLP}(z_t;\bm{\theta}_{v})$. In the meantime, we apply $M$ independent temporal convolutions with various kernel sizes in order to extract short-term patterns at different scales. Specifically, for $j=1,\ldots, M$, each convolution with parameter $\bm{\theta}_p$ takes the input $\rmX$ to give a sequence of local representations, $\rvp^j = \text{Conv}\left(\rmX;\bm{\theta}_p^j\right)$.
We then concatenate each $\rvp^j$ to build a multi-scale pattern embedding $\rmP = [\rvp^j]_{j=1}^M$ with parameters $\bm{\theta}_p = [\bm{\theta}_p^j]_{j=1}^M$ accordingly. To avoid dimension issues from the concatenation, we keep the dimension of $\rmP$ and $\rmV$ the same.
The extracted pattern $\rmP$ and value $\rmV$ are fed into the shared attention module.

\begin{figure*}[t]
    \centering
    \includegraphics[scale=0.6]{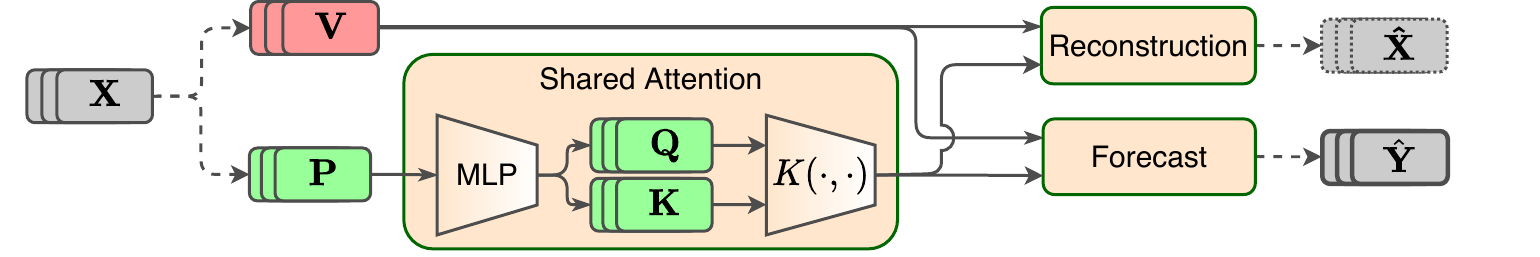}
    \caption{In DAF, the shared attention module processes pattern and value embeddings from either domain. A kernel function encodes pattern embeddings to a shared latent space for weight computation. We combine value embeddings by different groups of weights to obtain the interpolation $t\le T$ for reconstruction $\hat{\rmX}$ and the extrapolation $t=T+1$ for the forecast $\hat{\rmY}$.}
    \label{fig:attention} 
\end{figure*} 

\paragraph{Shared Attention Module}
We design the attention module to be shared by both domains since its primary task is to generate domain-invariant queries $\rmQ$ and keys $\rmK$ from pattern embeddings $\rmP$ for both source and target domains. 
Formally, we project $\rmP$ into $d$-dimensional queries $\rmQ=[\rvq_t]_{t=1}^{T}$ and keys $\rmK=[\rvk_t]_{t=1}^{T}$ via a position-wise MLP \[(\rvq_t, \rvk_t) = \text{MLP}(\rvp_t;\bm{\theta}_s).\]
As a result, the patterns from both domains are projected into a common space, which is later induced to be domain-invariant via adversarial training. At time $t$, an attention score $\alpha$ is computed as the normalized alignment between the query $\rvq_t$ and keys $\rvk_{t'}$ at neighborhood positions $t'\in\gN(t)$ using a positive semi-definite kernel $\gK(\cdot, \cdot)$,
\begin{equation}\label{eq:attn_score}
    \alpha(\rvq_t, \rvk_{t'}) = \frac{\gK(\rvq_t, \rvk_{t'})}{\sum_{t'\in \gN(t)}\gK(\rvq_t, \rvk_{t'})},
\end{equation}
e.g. an exponential scaled dot-product $\gK(\rvq, \rvk) = \exp\left(\frac{\rvq^T\rvk}{ \sqrt{d}}\right)$. 
Then, a representation $\rvo_t$ is produced as the average of values $\rvv_{\mu(t')}$ weighted by attention score $\alpha(\rvq_t, \rvk_{t'})$ on neighborhood $\mathcal{N}(t)$, followed by a MLP with parameter $\bm{\theta}_o$:
\begin{equation}\label{eq:attn_representation}
    \rvo_t = \text{MLP}\left(\sum_{t'\in \gN(t)}\alpha(\rvq_t, \rvk_{t'})\rvv_{\mu(t')};\bm{\theta}_o\right),
\end{equation}
where $\mu: \sN\rightarrow\sN$ is a position translation. The choice of $\gN(t)$ and $\mu(t)$ depends on whether $G$ is in interpolation mode for reconstruction when $t\leq T$ or extrapolation mode for forecasting when $t>T$. 

\begin{figure*}[h]
    \centering
    \includegraphics[scale=0.5]{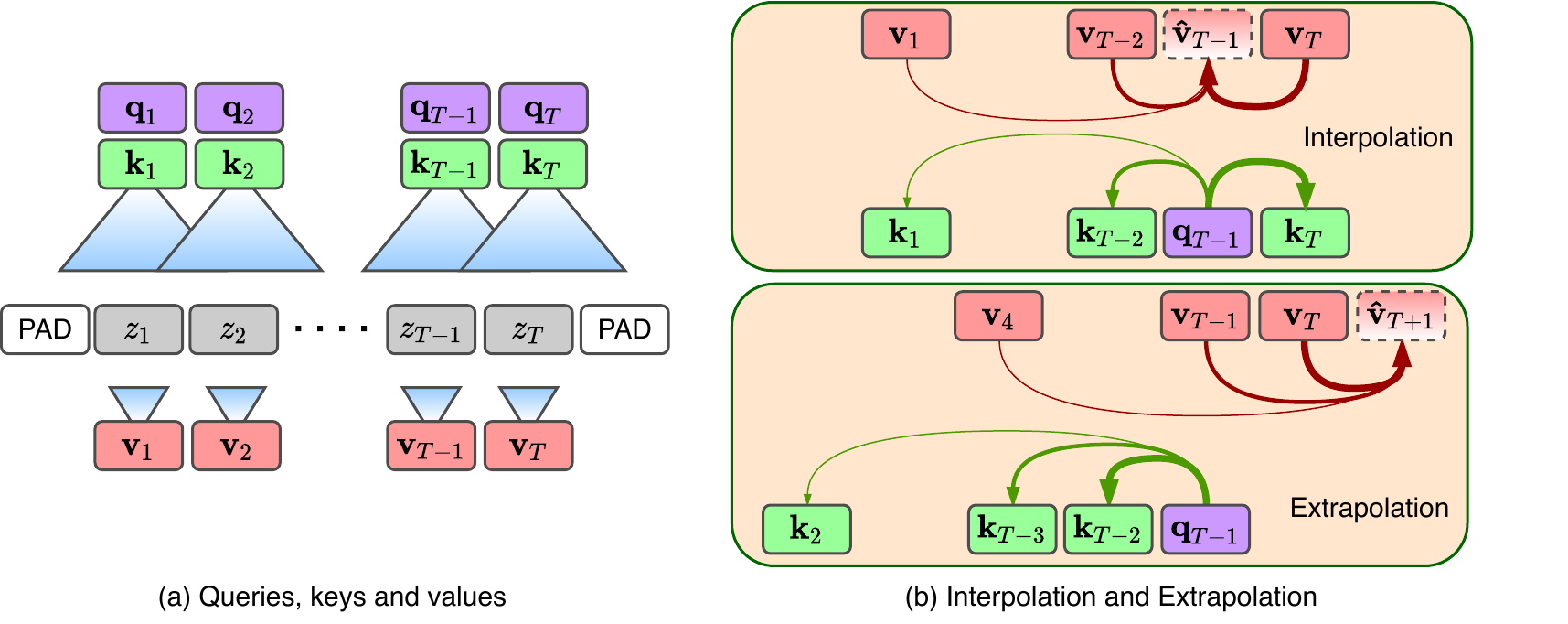}
    \caption{Interpolation and extrapolation within an attention module where $s=3$ and $M=1$. Specifically, the lower panel of (b) describes how attention is performed to estimate $\rvv_{T+1}$. The extrapolation query $q_{T+1}$ comes from the last local window $z_{T-2}, z_{T-1}, z_{T}$, which is convoluted into $q_{T-1}$. Meanwhile, a key $k_{t'}$ comes from the window $z_{t'-1}, z_{t'}, z_{t'+1}$ via convolution. Since the predicted $z_{T+1}$ follows the query window $z_{T-2}, z_{T-1}, z_{T}$, we take the value that follows the key window $z_{t'-1}, z_{t'}, z_{t'+1}$, i.e. $z_{t'+2}$, to make a correspondence. In this way, the following value of similar windows to the query window will receive larger weights in attention. For example, in Figure 7, the key $k_{T-2}$ corresponds to the value $v_T$, where the correspondence is indicated by the boldness of the arrows. Therefore, the key $k_{t'}$ will be paired with the value $v_{t'+\bar{s}+1}$, and $\mu(t') = t'+\bar{s}+1$. }
    \label{fig:attention}
\end{figure*}

\paragraph{Interpolation: Input Reconstruction}
We reconstruct the input by interpolating $\hat{z}_t$ using the observations at other time points. The upper panel of \Figref{fig:attention}(b) illustrates an example, where we would like to estimate $\hat{z}_{T-1}$ using $\{z_{1}, z_2, \dots, z_{T-2},z_{T}\}$. We take $\rvq_{T-1}$ that depends on the local windows centered at the target step $T-1$ as shown in \Figref{fig:attention}(a) as the query, and compare it with the keys $\{\rvk_{1}, \rvk_2, \dots, \rvk_{T-2},\rvk_{T}\}$. Similar to the query, the attended keys $\rvk_{t'}$ depend on local windows centered at the respective step $t'$. Hence, the scores $\alpha(\mathbf{q}_{T-1},\mathbf{k}_{t'})$, computed by \eqref{eq:attn_score} and illustrated by the thickness of arrows in \Figref{fig:attention}(b), depict the similarity of the value $\hat{z}_{T-1}$ to the attended value $z_{t'}$, and the output $\rvo_{T-1}$ is from the combination of $\rvv_{t'}$ weighted by $\alpha(\mathbf{q}_{T-1},\mathbf{k}_{t'})$ according to \eqref{eq:attn_score}.

By generalizing the example above at time $T-1$ to all time steps, we set
\begin{equation*}
\begin{aligned}
    \gN(t) &= \{1,2,\dots,T\} \setminus \{t\}, \\
    \mu(t') &= t',
\end{aligned}
\end{equation*}
in \eqref{eq:attn_representation}. 
Since the output $\rvo_t$ depends on values at $\gN(t)$ and does not access the ground truth $z_t$, the reconstruction task is not trivial.

\paragraph{Extrapolation: Future Predictions}
Since DAF is an autoregressive forecaster, it generates forecasts one step ahead. At each step, we forecast the next value by extrapolating from the given historical values. The lower panel of \Figref{fig:attention}(b) illustrates an example, where we would like to estimate the $(T+1)$-th value given the past $T$ observations and expected. The prediction $\hat{z}_{T+1}$ follows the last local window $\{z_{T-s+1}, z_{T-s+2}, \dots, z_{T}\}$ on which the query $\rvq_{T-\bar{s}}$ is dependent, where $\bar{s} = \ceil{\frac{s-1}{2}}$, and $\ceil{\cdot}$ denotes the ceiling operator. We take $\rvq_{T-\bar{s}}$ as the query for $T+1$, i.e. we set $\rvq_{T+1} = 
\rvq_{T-\bar{s}}$, and attend to the previous keys that do not encode padding zeros, i.e. we set:
\[
    \gN(T+1)=\{s,\dots,T-\bar{s}-1\}.
\]
In this case, the attention score $\alpha(\mathbf{q}_{T+1},\mathbf{k}_{t'})$ from \eqref{eq:attn_score} depicts the similarity of the unknown $\hat{z}_{T+1}$, and the value $z_{t'+\bar{s}+1}$ following the local window $\{z_{t'-\bar{s}}, \dots, z_{t'}, \dots, z_{t'+\bar{s}}\}$ corresponding to the attended key $\rvk_{t'}$. Hence, we set 
\[
    \mu(t')=t'+\bar{s}+1,
\]
in \eqref{eq:attn_representation} to estimate $\rvo_{T+1}$.

\Figref{fig:attention} illustrates an example of future forecasts, where $s=3$ and $M=1$ in encoder module. A detailed walkthrough of can be found in the caption.


\paragraph{Private Decoders}
The private decoder produces prediction $\hat{z}_t$ out of $\rvo_t$ through another position-wise MLP: $\hat{z}_t = \text{MLP}(\rvo_t;\bm{\theta}_d)$.
By doing so, we can generate reconstructions $\hat{\rmX} = [\hat{z}_t]_{t=1}^{T}$ and the one-step prediction $\hat{z}_{T+1}$ . This prediction $\hat{z}_{T+1}$ is fed back into the encoder and attention model to predict the next one-step ahead prediction. We recursively feed the prior predictions to generate the predictions $\hat{\rmY}=[\hat{z_t}]_{t=T+1}^{T+\tau}$ over $\tau$ time steps.

\subsection{Domain Discriminator}\label{sec:adv}
In order to induce the queries and keys of the attention module to be domain-invariant, a domain discriminator is introduced to recognize the origin of a given query or key. A position-wise MLP-based binary classifier $D: \sR^d \rightarrow [0,1]$:
\[
D(\rvq_t) = \text{MLP}( \rvq_t;\bm{\theta}_D),
~
D(\rvk_t) = \text{MLP}( \rvk_t;\bm{\theta}_D).
\]
is trained by minimizing the cross entropy loss of $\gL_{dom}$ in \eqref{eq:cross_entropy}. We design the latent features $\mathcal{H_{\gS}}, \mathcal{H_{\gT}}$ in \eqref{eq:cross_entropy} to be the keys $\rmK=[\rvk_t]_{t=1}^{T+\tau}$ and queries $\rmQ=[\rvq_t]_{t=1}^{T+\tau}$ in both source and target domains, respectively. 

\subsection{Adversarial Training}\label{sec:at}
\begin{algorithm}[tb]
\caption{Adversarial Training of DAF}
\begin{algorithmic}[1]\label{alg:example}
    \STATE \textbf{Input:} dataset $\gD_\gS$, $\gD_\gT$; epochs $E$, step sizes
    \STATE \textbf{Initialization}: parameter $\bm{\Theta}_G$ for generator $G_\gS$, $G_\gT$, parameter $\bm{\theta}_D$ for discriminator $D$
    \FOR{epoch $=1$ {\bfseries to} $E$}
    \REPEAT
        \STATE sample $\rmX_\gS, \rmY_\gS \sim \gD_\gS$ and $\rmX_\gT,\rmY_\gT \sim \gD_\gT$
        \STATE generate $\hat{\rmX}_\gS, \hat{\rmY}_\gS = G_\gS(\rmX_\gS)$
        and $\hat{\rmX}_\gT, \hat{\rmY}_\gT = G_\gT(\rmX_\gT)$
        \STATE compute  $\gL_{seq}$ in \eqref{eq:l_seq} for $\gS$ and $\gT$
        , $\gL_{dom}$ in  \eqref{eq:cross_entropy}
        , and total $\gL$ in \eqref{eq:minimax_obj}
        \STATE gradient descent with $\nabla_{\bm{\Theta}_G}\gL$ to update $G_\gS$, $G_\gT$
        \STATE gradient ascent with $\nabla_{\bm{\theta}_D}\gL$ to update $D$
    \UNTIL $\gD_\gT$ is exhausted
    \ENDFOR
\end{algorithmic}
\end{algorithm}

Recall we have defined generators $ G_\gS, G_\gT$ based on the private encoder/decoder and the shared attention module. The discriminator $D$ induces the invariance of latent features keys $\rmK$ and queries $\rmQ$ across domains. While $D$ tries to classify the domain between source and target, $G_\gS, G_\gT$ are trained to confuse $D$. 
By choosing the MSE loss for $l$, the minimax objective in \eqref{eq:minimax_obj} is now formally defined over generators $G_\gS, G_\gT$ with parameters $\bm{\Theta}_G = \{ \bm{\theta}_p^\gS, \bm{\theta}_v^\gS,\bm{\theta}_d^\gS, \bm{\theta}_p^\gT, \bm{\theta}_v^\gT,\bm{\theta}_d^\gT, \bm{\theta}_s, \bm{\theta}_o\}$ and domain discriminator $D$ with parameter $\bm{\theta}_D$. 
Algorithm \ref{alg:example} summarizes the training routine of DAF. We alternately update $\bm{\Theta}_G$ and $\bm{\theta}_D$ in opposite directions so that $G=\{G_\gS, G_\gT\}$ and $D$ are trained adversarially. Here, we use a standard pre-processing for $\rmX, \rmY$ and post-processing for $\hat \rmX, \hat \rmY$. In our experiments, the coefficient $\lambda$ in \eqref{eq:minimax_obj} is fixed to be $1$.


\section{Experiments}\label{sec:exp}
We conduct extensive experiments to demonstrate the effectiveness of the proposed DAF in adapting from a source domain to a target domain, leading to accuracy improvement over state-of-the-art forecasters and existing DA ethos. In addition, we conduct ablation studies to examine the contribution of our design to the significant performance improvement. 

\subsection{Baselines and Evaluation}\label{sec:baseline}
In the experiments, we compare DAF with the following single-domain and cross-domain baselines.  
The conventional single-domain forecasters trained only on the target domain include:
\begin{itemize}[noitemsep,topsep=0pt, leftmargin=1.5em]
    \item DAR: DeepAR \citep{flunkert_deepar_2020};
    \item VT: Vanilla Transformer \citep{vaswani_attention_2017};
    \item AttF: the sequence generator $G_{\gT}$ for the target domain trained by minimizing $\gL_{seq}(\gD_{\gT};G_{\gT})$ in \eqref{eq:minimax_obj}. 
\end{itemize}
The cross-domain forecasters trained on both source and target domain include:
\begin{itemize}[noitemsep,topsep=0pt, leftmargin=1.5em]
    \item DATSING: pretrained and finetuned forecaster \cite{hu_datsing_2020};
    \item SASA: metric-based domain adaptation for time series data \cite{cai_time_2021} which is extended from regression task to multi-horizon forecasting;
    \item RDA: RNN-based DA forecaster obtained by replacing the attention module in DAF with a LSTM module and inducing the domain-invariance of LSTM encodings. Specifically, we consider three variants:
    \begin{itemize}
        \item RDA-DANN: adversarial DA via gradient reversing \cite{ganin_domain-adversarial_2016};
        \item RDA-ADDA: adversarial DA via GAN-like optimization \cite{tzeng_adversarial_2017};
        \item RDA-MMD: metric based DA via minimizing MMD between LSTM encodings \cite{li_mmd_2017}.
    \end{itemize}
\end{itemize}
We implement the models using PyTorch \citep{paszke_pytorch_2019}, and train them on AWS Sagemaker \citep{liberty_elastic_2020}. For DAR, we call the publicly available version on Sagemaker. In most of the experiments, DAF and the baselines are tuned on a held-out validation set.  See appendix \ref{sec:hyperp} for details on the model configurations and hyperparameter selections.

We evaluate the forecasting error in terms of the Normalized Deviation (ND) 
\citep{ yu_temporal_2016}:
$$\text{ND} = \left(\sum_{i=1}^N\sum_{t=T+1}^{T+\tau}\vert z_{i,t}-\hat{z}_{i,t}\vert\right)/\left(\sum_{i=1}^N\sum_{t=T+1}^{T+\tau}\vert z_{i,t}\vert\right),$$
where $\rmY_i = [z_{i,t}]_{t=T+1}^{T+\tau}$ and $\hat{\rmY_i} = [\hat{z}_{i,t}]_{t=T+1}^{T+\tau}$ denote the ground truths and predictions, respectively.  In the subsequent tables, the methods with a mean ND metric within one standard deviation of method with the lowest mean ND metric are shown in bold. 

\subsection{Synthetic Datasets}\label{sec:synthetic}

\begin{table*}[h]

\centering
\scalebox{0.95}{
    \begin{tabular}{c|c|c|c|ccc|cc|c}
    \toprule
    Task                    & $N$ & $T$& $\tau$ &DAR                & VT          & AttF       & DATSING  & RDA-ADDA                       & DAF                      \\ \midrule
    \multirow{3}{*}{\begin{tabular}[c]{@{}c@{}}Cold\\Start\end{tabular}} &  \multirow{3}{*}{5000} &36  &\multirow{6}{*}{18} & 0.053$\pm$0.003       & 0.040$\pm$0.001 & 0.042$\pm$0.001 & 0.039$\pm$0.004 & \textbf{0.035$\pm$0.002} & \textbf{0.035$\pm$0.003} \\ \cline{3-3} \cline{5-10}
                                   &&45&  & 0.037$\pm$0.002       & 0.039$\pm$0.001 & 0.041$\pm$0.004 & 0.039$\pm$0.002 &  0.034$\pm$0.001    & \textbf{0.030$\pm$0.003} \\ \cline{3-3} \cline{5-10}
                                & &54  & & \textbf{0.031$\pm$0.002} & 0.039$\pm$0.001 & 0.038$\pm$0.005 &  0.037$\pm$0.001       & 0.034$\pm$0.001          & \textbf{0.029$\pm$0.003} \\ \cline{1-3}\cline{5-10}
    \multirow{3}{*}{\begin{tabular}[c]{@{}c@{}}Few\\Shot\end{tabular}}  & 20 &\multirow{3}{*}{144} &  & 0.062$\pm$0.003       & 0.089$\pm$0.001 & 0.095$\pm$0.003 &  0.078$\pm$0.005 & \textbf{0.059$\pm$0.003}   & \textbf{0.057$\pm$0.004} \\
                                \cline{2-2}\cline{5-10}
                                 &  50 & &  & 0.059$\pm$0.004       & 0.085$\pm$0.001 & 0.074$\pm$0.005 &  0.076$\pm$0.006        & \textbf{0.054$\pm$0.003} &  \textbf{0.055$\pm$0.001}   \\ 
                                \cline{2-2}\cline{5-10} 
                                 &  100 & &  & 0.059$\pm$0.003       & 0.079$\pm$0.002 & 0.071$\pm$0.002 & 0.058$\pm$0.005   &  0.053$\pm$0.007    & \textbf{0.051$\pm$0.001} \\ \bottomrule
    \end{tabular}
}
\caption{
Performance comparison of DAF on synthetic datasets with varying historical lengths $T$ (cold-start), and varying number of time series $N$ (few-shot) and prediction length $\tau$ in terms of the mean +/- the standard deviation ND metric. The winners and the competitive followers (the gap is smaller than its standard deviation over 5 runs) are bolded for reference. \vspace{-.4cm}
}
\label{tab:synthetic}
\end{table*}


We first simulate scenarios suited for domain adaptation, namely \textbf{cold-start} and \textbf{few-shot} forecasting. In both scenarios, we consider a source dataset $\gD_{\gS}$ and a target dataset $\gD_{\gT}$ consisting of time-indexed sinusoidal signals with random parameters, including amplitude, frequency and phases, sampled from different uniform distributions. See appendix \ref{sec:dataset} for details on the data generation. The total observations in the target dataset are limited in both scenarios by either length or number of time series.

\textbf{Cold-start} forecasting aims to forecast in a target domain, where the signals are fairly short and limited historical information is available for future predictions. To simulate solving the cold-start problem, we set the time series historical length in the source data $T_{\gS}=144$, and vary the historical length in the target data $T$ within $\{36, 45, 54\}$. The period of sinusoids in the target domain is fixed to be $36$, so that the historical observations cover $1\sim1.5$ periods. We also fix the number of time series $N_{\gS} = N = 5000$.

\textbf{Few-shot} forecasting occurs when there is an insufficient number of time series in the target domain for a well-trained forecaster. To simulate this problem, we set the number of time series in the source data $N_{\gS} = 5000$, and vary the number of time series in the target data $N$ within $\{20, 50, 100\}$.  We also fix the historical lengths $T_{\gS}=T=144$.
The prediction length is set to be equal for both source and target datasets, i.e. $\tau_{\gS}=\tau=18$. 

\begin{figure}[H]
    \centering
    \includegraphics[scale=0.3]{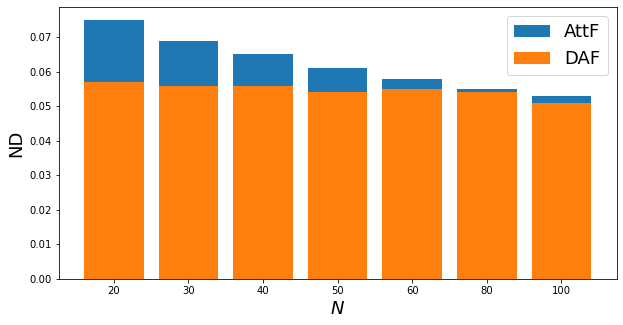}
    \caption{
        Forecasting accuracy of AttF and DAF methods in synthetic few-shot experiments with different target dataset sizes.
        \vspace{-.2cm}
    }
    \label{fig:improvement}
\end{figure}

The results of the synthetic experiments on the cold-start and few-shot problems in \Tabref{tab:synthetic} demonstrate that the performance of DAF is better than or on par with the baselines in all experiments.  
We also note the following observations to provide a better understanding into domain adaptation methods.  
First, we see that the cross-domain forecasters RDA and DAF that are jointly trained end-to-end using both source and target data are overall more accurate than the single-domain forecasters. 
This finding indicates that source data is helpful in forecasting the target data.
Second, among the cross-domain forecasters DATSING is outperformed by RDA and DAF, indicating the importance of joint training on both domains.
Third, on a majority of the experiments our attention-based DAF model is more accurate than or competitive to the RNN-based DA (RDA) method. We show the results for RDA-ADDA as the other DA variants, DANN and MMD, have similar performance. They are considered in the following real-world experiments (see \Tabref{tab:real}).
Finally, we observe in \Figref{fig:improvement} that DAF improves more significantly as the number of training samples becomes smaller.

\subsection{Real-World Datasets}
\label{subsec:real}

\begin{table*}[h]
\centering
\scalebox{0.8}{
    \begin{tabular}{c|c|c|cc|ccccc|c}
    \toprule
    $\gD_{\gT}$            & $\gD_{\gS}$    & $\tau$              & DAR                              & AttF                             & SASA            & DATSING         & RDA-DANN        & RDA-ADDA                 & RDA-MMD                  & DAF                      \\ \midrule
    \multirow{2}{*}{traf}  & \textit{elec}  & \multirow{4}{*}{24} & \multirow{2}{*}{0.205$\pm$0.015} & \multirow{2}{*}{0.182$\pm$0.007} & 0.177$\pm$0.004 & 0.184$\pm$0.004 & 0.181$\pm$0.009 & 0.174$\pm$0.005          & 0.186$\pm$0.004          & \textbf{0.169$\pm$0.002} \\ \cline{2-2}
                           & \textit{wiki}  &                     &                                  &                                  & 0.197$\pm$0.001 & 0.189$\pm$0.005 & 0.180$\pm$0.004 & 0.181$\pm$0.003          & 0.179$\pm$0.004          & \textbf{0.176$\pm$0.004} \\ \cline{1-2}\cline{4-11}
    \multirow{2}{*}{elec}  & \textit{traf}  &                     & \multirow{2}{*}{0.141$\pm$0.023} & \multirow{2}{*}{0.137$\pm$0.005} & 0.164$\pm$0.001 & 0.137$\pm$0.003 & 0.133$\pm$0.005 & 0.134$\pm$0.002          & 0.140$\pm$0.006          & \textbf{0.125$\pm$0.008} \\ \cline{2-2}
                           & \textit{sales} &                     &                                  &                                  & 0.160$\pm$0.001 & 0.149$\pm$0.009 & 0.135$\pm$0.007 & 0.142$\pm$0.003          & 0.144$\pm$0.003          & \textbf{0.123$\pm$0.005} \\ \midrule
    \multirow{2}{*}{wiki}  & \textit{traf}  & \multirow{4}{*}{7}  & \multirow{2}{*}{0.055$\pm$0.010} & \multirow{2}{*}{0.050$\pm$0.003} & 0.053$\pm$0.001 & 0.049$\pm$0.002 & 0.047$\pm$0.005 & \textbf{0.045$\pm$0.003} & \textbf{0.045$\pm$0.003} & \textbf{0.042$\pm$0.004} \\ \cline{2-2}
                           & \textit{sales} &                     &                                  &                                  & 0.053$\pm$0.001 & 0.052$\pm$0.004 & 0.053$\pm$0.002 & \textbf{0.049$\pm$0.003} & 0.052$\pm$0.004          & \textbf{0.049$\pm$0.003} \\ \cline{1-2}\cline{4-11}
    \multirow{2}{*}{sales} & \textit{elec}  &                     & \multirow{2}{*}{0.305$\pm$0.005} & \multirow{2}{*}{0.308$\pm$0.002} & 0.451$\pm$0.001 & 0.301$\pm$0.008 & 0.297$\pm$0.004 & \textbf{0.281$\pm$0.001} & 0.291$\pm$0.004          & \textbf{0.277$\pm$0.005} \\ \cline{2-2}
                           & \textit{wiki}  &                     &                                  &                                  & 0.301$\pm$0.001 & 0.305$\pm$0.008 & 0.287$\pm$0.009 & \textbf{0.287$\pm$0.002} & 0.289$\pm$0.003          & \textbf{0.280$\pm$0.007} \\
    \bottomrule
    \end{tabular}
}
\caption{Performance comparison of DAF on real-world benchmark datasets with 
prediction length $\tau$ in the target domain in terms of the mean +/- standard deviation ND metric. The winners and the competitive followers (the gap is smaller than its standard deviation over 5 runs) are bolded for reference.\vspace{-.2cm}} 
\label{tab:real}
\end{table*}



\begin{figure}[h]
    \centering
    \includegraphics[scale=0.4]{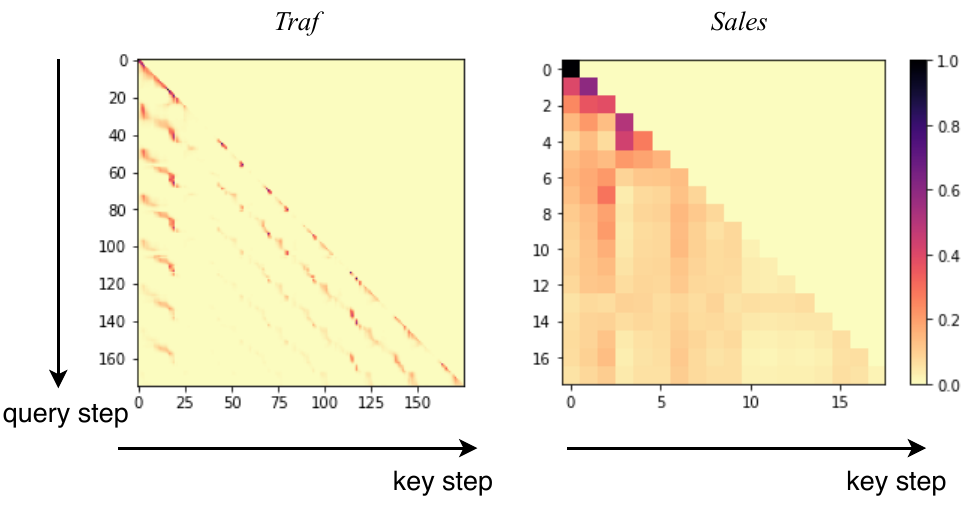}
    \caption{Attention distribution produced by an attention head of DAF with $\gD_{\gS} = $ \textit{traf} (left) and $\gD_{\gT} =$ \textit{sales} (right).}
    \label{fig:heatmap}
\end{figure}

We perform experiments on four real benchmark datasets that are widely used in forecasting literature: \textit{elec} and \textit{traf} from the UCI data repository \citep{dua_uci_2017}, \textit{sales} \citep{kar_dataset_2019} and \textit{wiki} \citep{lai_dataset_2017} from Kaggle. Notably, the \textit{elec} and \textit{traf} datasets present clear daily and weekly patterns while \textit{sales} and \textit{wiki} are less regular and more challenging. We use the following time features $\xi_t \in \mathbb{R}^2$ as covariates: the day of the week and hour of the day for the hourly datasets \textit{elec} and \textit{traf}, and the day of the month and day of the week for the daily datasets \textit{sales} and \textit{wiki}. For more dataset details, see appendix \ref{sec:dataset}. 

To evaluate the performance of DAF, we consider cross-dataset adaptation, i.e., transferring between a pair of datasets. Since the original datasets are large enough to train a reasonably good forecaster, we only take a subset of each dataset as a target domain to simulate the data-scarce situation. Specifically, we take the last $30$ days of each time series in the hourly dataset \textit{elec} and \textit{traf}, and the last $60$ days from daily dataset \textit{sales} and \textit{wiki}. We partition the target datasets equally into training/validation/test splits, i.e. $10/10/10$ days for hourly datasets and $20/20/20$ days for daily datasets. The full datasets are used as source domains in adaptation. We follow the rolling window strategy from \citet{flunkert_deepar_2020}, and split each window into historical and prediction time series of lengths $T$ and $T + \tau$, respectively. In our experiments, we set $T=168, \tau=24$ for hourly datasets, and $T=28, \tau=7$ for the daily datasets. For DA methods, the splitting of the source data follows analogously.

Table \ref{tab:real} shows that the conclusions drawn from Table \ref{tab:synthetic} on the synthetic experiments generally hold on the real-world datasets.  In particular, we see the accuracy improvement by DAF over the baselines is more significant than that in the synthetic experiments.  The real-world experiments also demonstrate that in general the success of DAF is agnostic of the source domain, and is even effective when transferring from a source domain of different frequency than that of the target domain. In addition, the cross-domain forecasters, DATSING, the RDA variants and our DAF outperform the three single-domain baselines in most cases.
As in the synthetic cases, DATSING performs relatively worse than RDA and DAF. On the other hand, another cross-domain forecaster SASA is originally designed for regression tasks, where a fixed-sized window is used to predict a single exogeneous numerical label. While it can be extended to multi-horizon forecasting tasks by replacing the label with the next-step value and making autoregressive predictions, the performance is significantly worse than other cross-domain competitors and even some single-domain counterparts in some cases.
The accuracy differences between DAF and RDA are larger than in the synthetic case, and in favor of DAF.  This finding further demonstrates that our choice of an attention-based architecture is well-suited for real domain adaptation problems.

Remarkably, DAF manages to learn the different patterns between source and target domains under our setups. For instance, \Figref{fig:heatmap} illustrates that DAF can successfully learn clear daily patterns in the \textit{traf} dataset, and find irregular patterns in the \textit{sales} dataset. A reason for its success is that the private encoders capture features at various scales in different domains, and the attention module captures domain-dependent patterns by context matching using domain-invariant queries and keys.

\begin{table}[h]
    \centering
    \scalebox{0.75}{
    \begin{tabular}{c|c|c|ccccc}
    \toprule
    $\gD_\gT$ & $\gD_\gS$ & $\tau$ & \textit{no-adv} & \textit{no-$q$-share} & \textit{no-$k$-share} & \textit{$v$-share} & DAF            \\ \hline
    \textit{traf}      & \textit{elec} &24     & 0.172  & 0.171        & 0.172        & 0.176     & \textbf{0.168} \\ \hline
    \textit{elec}      & \textit{traf}  &24    & 0.121  & 0.122        & 0.120        & 0.127     & \textbf{0.119} \\ \hline
    \textit{wiki}     & \textit{sales}   &7   & 0.042  & 0.042        & 0.044        & 0.049     & \textbf{0.041} \\ \hline
    \textit{sales}      & \textit{wiki} &7     & 0.294  & 0.283        & 0.282        & 0.291     & \textbf{0.280} \\ \bottomrule
    \end{tabular}}
    \caption{Results of ablation studies of DAF variants on four adaptation tasks on real-world datasets.\vspace{-.2cm}}
    \label{tab:abl}
\end{table}

\subsection{Additional Experiments}
In addition to the listed baselines in \secref{sec:baseline}, we also compare DAF with other single-domain forecasters, e.g. ConvTrans \cite{li_enhancing_2019}, N-BEATS \cite{oreshkin_n-beats_2020}, and domain adaptation methods on time series tasks, e.g. MetaF \cite{oreshkin_meta-learning_2020}. These methods are either similar to the baselines in \Tabref{tab:real} or designed for a different setting. We still adapt them to our setting to provide additional results in Tables \ref{tab:full_syn_result}-\ref{tab:full_real_result} in appendix \ref{sec:dr}. 

\subsection{Ablation Studies}
In order to examine the effectiveness of our designs, we conduct ablation studies by adjusting each key component successively. 
\Tabref{tab:abl} shows the improved performance of DAF over its variants on the target domain on four adaptation tasks.  
Equipped with a domain discriminator, DAF improves its effectiveness of adaptation compared to its non-adversarial variant (\textit{no-adv}). 
We see that sharing both keys and queries in DAF results in performance gains over not sharing either (\textit{no-$k$-share} and \textit{no-$q$-share}). Furthermore, it is clear that our design choice of the values to be domain-specific for domain-dependent forecasts rather than shared (\textit{$v$-share}) has the largest positive impact on the performance.

\begin{figure}[h]
 \centering
    \includegraphics[scale=0.45]{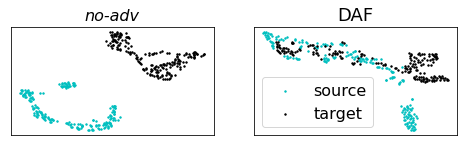}
    \captionof{figure}{Query alignment with (DAF: right) and without adversarial training (\textit{no-adv}: left), where $\gD_{\gS} =$ \textit{elec} and $\gD_{\gT} =$ \textit{traf}.\vspace{-.6cm}}
    \label{fig:alignment}
\end{figure}

Figure \ref{fig:alignment} visualizes the distribution of queries and keys learned by DAF and \textit{no-adv}, where the target data $\gD_{\gT} = \textit{traf}$ and the source data $\gD_{\gS} = \textit{elec}$ via a TSNE embedding. Empirically, we see the latent distributions are well aligned in DAF and not in \textit{no-adv}. This can explain the improved performance of DAF over its variants. It also further verifies our intuition that DAF benefits from an aligned latent space of queries and keys across domains. 


\section{Conclusions} \label{sec:conc}
In this paper, we aim to apply domain adaptation to time series forecasting to solve the data scarcity problem. We identify the differences between the forecasting task and common domain adaptation scenarios, and accordingly propose the Domain Adaptation Forecaster (DAF) based on attention sharing. Through empirical experiments, we demonstrate that DAF outperforms state-of-the-art single-domain forecasters and various domain adaptation baselines on synthetic and real-world datasets.  We further show the effectiveness of our designs via extensive ablation studies.
In spite of empirical evidences, the theoretical justification of having domain-invariant features within attention models remains an open problem.  
Extension to multi-variate time series forecasting experiments is another direction of future work.

\bibliography{references.bib}
\bibliographystyle{icml2022}

\include{appendix}
\end{document}

%% file: math_commands.tex
\usepackage{amsmath,amsfonts,bm}

\def\Figref#1{Figure~\ref{#1}}

\def\Tabref#1{Table~\ref{#1}}

\def\secref#1{section~\ref{#1}}

\def\eqref#1{equation~(\ref{#1})}

\def\ceil#1{\lceil #1 \rceil}

\def\1{\bm{1}}

\def\rvk{{\mathbf{k}}}

\def\rvo{{\mathbf{o}}}
\def\rvp{{\mathbf{p}}}
\def\rvq{{\mathbf{q}}}

\def\rvv{{\mathbf{v}}}

\def\rmK{{\mathbf{K}}}

\def\rmP{{\mathbf{P}}}
\def\rmQ{{\mathbf{Q}}}

\def\rmV{{\mathbf{V}}}

\def\rmX{{\mathbf{X}}}
\def\rmY{{\mathbf{Y}}}

\DeclareMathAlphabet{\mathsfit}{\encodingdefault}{\sfdefault}{m}{sl}
\SetMathAlphabet{\mathsfit}{bold}{\encodingdefault}{\sfdefault}{bx}{n}

\def\gD{{\mathcal{D}}}

\def\gH{{\mathcal{H}}}

\def\gK{{\mathcal{K}}}
\def\gL{{\mathcal{L}}}

\def\gN{{\mathcal{N}}}

\def\gS{{\mathcal{S}}}
\def\gT{{\mathcal{T}}}

\def\sN{{\mathbb{N}}}

\def\sR{{\mathbb{R}}}



%% file: appendix.tex
\newpage

\appendix




\newpage

\section{Dataset Details}\label{sec:dataset}
\subsection{Synthetic Datasets}
The synthetic datasets consist of sinusoidal signals with uniformly sampled parameters as follows:
\begin{equation*}
\begin{aligned}
    z_{i,t} &= A_i\sin(2\pi\omega_it+\phi_i) + c_i + \epsilon_{i,t}, \quad t \in [0, T+\tau], \\
    A_i &\sim \text{Unif}(A_{\min}, A_{\max}),\quad c_i \sim \text{Unif}(c_{\min}, c_{\max}), \\
    \omega_i &\sim \text{Unif}(\omega_{\min}, \omega_{\max}),\quad \phi_i \sim \text{Unif}(-2\pi, 2\pi), \\
\end{aligned}
\end{equation*}     
where $A_{\min}, A_{\max} \in \mathbb{R}^+$ denote the amplitudes, $c_{\min}, c_{\max}\in\mathbb{R}$ denote the levels, and $\omega_{\min}, \omega_{\max}\in [\frac{1}{T}, \frac{20}{T}]$ denote the frequencies. In addition, $\epsilon_{i,t}\sim\gN(0,0.2)$ is a white noise term. In our experiments, we fix $T = 144, \tau=18, A_{\min} = 0.5, A_{\max}=5.0$, $c_{\min} = -3.0, c_{\max} = 3.0$.

\subsection{Real-World Datasets}
\Tabref{tab:stats} summarizes the four benchmark real-world datasets that we use to evaluate our DAF model. 
\begin{table}[h]

\centering
    \scalebox{0.85}{
    \begin{tabular}{|c|c|c|c|c|l|}
    \hline
    Dataset & Freq   & Value          & \begin{tabular}[c]{@{}c@{}}\# Time \\ Series\end{tabular} & \begin{tabular}[c]{@{}c@{}}Average \\ Length\end{tabular} & Comment                                                                                  \\ \hline
    \textit{elec}    & hourly & $\mathbb{R}^+$ & 370                                                       & 3304                                                      & \begin{tabular}[c]{@{}l@{}}Household \\ electricity\\ consumption\end{tabular}        \\ \hline
    \textit{traf}    & hourly & $[0,1]$        & 963                                                       & 360                                                       & \begin{tabular}[c]{@{}l@{}}Occupancy rate \\ of SF Bay Area \\ highways\end{tabular}  \\ \hline
    \textit{sales}   & daily  & $\mathbb{N}^+$ & 500                                                       & 1106                                                      & \begin{tabular}[c]{@{}l@{}}Daily sales of \\ Rossmann\\ grocery stores\end{tabular}   \\ \hline
    \textit{wiki}    & daily  & $\mathbb{N}^+$ & 9906                                                      & 70                                                        & \begin{tabular}[c]{@{}l@{}}Visit counts of \\ various Wikipedia\\  pages\end{tabular} \\ \hline
    \end{tabular}}
\caption{Benchmark dataset descriptions.}
\label{tab:stats}
\end{table}

For evaluation, we follow \citet{flunkert_deepar_2020}, and we take moving windows of length $T+\tau$ starting at different points from the original time series in the datasets. For an original time series $[z_{i,t}]_{t=1}^L$ of length $L$, we obtain a set of moving windows: $$\{[z_{i,t}]_{t=n}^{n+T+\tau-1}, \qquad n=1,2,\dots,L-T-\tau+1\}.$$ This procedure results in a set of fixed-length trajectory samples. Each sample is further split into historical observations $X$ and forecasting targets $Y$, where the lengths of $X$ and $Y$ are $T$ and $\tau$, respectively. We randomly select samples from the population by uniform sampling for training, validation and test sets.

\section{Implementation Details}\label{supp:impl_detail}
\subsection{Baselines}
In this subsection, we provide an overview of the following baseline models, including
conventional single-domain forecasters trained only on the target domain:
\begin{itemize}
    \item DeepAR (DAR): auto-regressive RNN-based model with LSTM units (we directly call DeepAR implemented by Amazon Sagemaker);
    \item Vanilla Transformer (VT): sequence-to-sequence model based on common transformer architecture;
    \item ConvTrans (CT): attention-based forecaster that builds attention blocks on convolutional activations as DAF does. Unlike DAF, it does not reconstruct the input, and only fits the future. From a probabilistic perspective, it models the conditional distribution $P(Y|X)$ instead of the joint distribution $P(X,Y)$ as DAF does, where $X$ and $Y$ are history and future, respective. In addition, it directly uses the outputs of the convolution as queries and keys in the attention module;
    \item AttF: single-domain version of DAF. It is equivalent to the branch of the sequence generator for the target domain in DAF. It has access to the attention module, but does not share it with another branch;
    \item N-BEATS: MLP-based forecaster. Similar to DAF, it aims to forecast the future as well as to reconstruct the given history. N-BEATS only consumes univariate time series $z_{i,t}$, and does not accept covariates $\xi_{i,t}$ as input. In the original paper \citet{oreshkin_n-beats_2020}, it employs various objectives and ensembles to improve the results. In our implementation, we use the ND metric as the training objective, and do not use any ensembling techniques for a fair comparison;
\end{itemize}
and cross-domain forecasters trained on both source and target domain:
\begin{itemize}
    \item MetaF: method with the same architecture as N-BEATS. It trains a model on the source dataset, and applies it to the target dataset in a zero-shot setting, i.e. without any fine-tuning. Both MetaF and N-BEATS rely on given Fourier bases to fit seasonal patterns. For instance, bases of period 24 and 168 are included for hourly datasets, whereas bases of period 7 are included for daily datasets. The daily and weekly patterns are expected to be captured in all settings. 
    \item PreTrained Forecaster (PTF): method with the same architecture as AttF for the target data. Unlike AttF, PTF fits both source and target data. It is first pretrained on the source dataset, and then finetuned on the target dataset. 
    \item SASA: \citet{cai_time_2021} is another related work that we introduce in \secref{sec:liter}, which focuses on time series classification and regression tasks, where a single exogenous label instead of a sequence of future values is predicted. We replace the original labels of the regression task with the next value of the respective time series, and autoregressively predicting the next step for multi-horizon forecasting. In the experiments, we revise the official code \footnote{https://github.com/DMIRLAB-Group/SASA.} accordingly and keep the provided default hyperparameters. 
    \item DATSING: a forecasting framework based on NBEATS \cite{oreshkin_n-beats_2020} architecture. The model is first pre-trained on the source dataset. Then a subset of source data that includes nearest neighbors to a target sample in terms of soft-DTW \cite{cuturi_soft-dtw:_2017} is selected to fine-tune the pre-trained model before it is evaluated using the respective target sample. During fine-tuning, a domain discriminator is used to distinguish the nearest neighbors from the same number of source samples drawn from the complement set.
    \item RDA has the same overall structure as DAF, but replaces the attention module in DAF with a LSTM module. The encoder module produces a single encoding, which is then consumed by the MLP-based decoder. We take three traditional methods for domain adaptation:
    \begin{itemize}
        \item RDA-DANN: The encoder and decoder are shared across domains. The sequence generator is trained to fit data from both domains. Meanwhile, the gradient of domain discriminator will be reversed before back-propagated to the encoder.
        \item RDA-ADDF: The encoder and decoder are not shared across domains. During training, the source encoder and decoder are first trained to fit source data. Then encodings from both encoders are discriminated by the domain discriminator with the source encoder parameters frozen. Finally, the target encoder and decoder are trained to fit target data with the target encoder parameters frozen.
        \item RDA-MMD: Instead of a domain discriminators, the Maximum Mean Discrepancy between source and target encodings is optimized with the sequence generators.
    \end{itemize}
\end{itemize}

\subsection{Hyperparameters}\label{sec:hyperp}
The following hyperparameters of DAF and baseline models are selected by grid-search over the validation set: 
\begin{itemize}
    \item the hidden dimension $h \in \{32, 64, 128, 256\}$ of all models;
    \item the number of MLP layers $l_{\text{MLP}} \in \{4\}$ for N-BEATS \footnote{We set the other hyperparameters for N-BEATS as suggested in the original paper \citet{oreshkin_n-beats_2020}.}, $l_{\text{MLP}} \in \{1,2 ,3\}$ for AttF, DAF and its variants;
    \item the number of RNN layers $l_{\text{RNN}} \in \{1,3\}$ in DAR and RDA;
    \item the kernel sizes of convolutions $s \in \{3,13,(3,5),(3,17)\}$ in AttF, DAF and its variants;
    \footnote{A single integer means a single convolution layer in the encoder module, while a tuple stands for multiple convolutions.}
    \item the learning rate $\gamma \in \{0.001, 0.01, 0.1\}$ for all models;
    \item the trade-off coefficient $\lambda \in \{0.1, 1, 10\}$ in \eqref{eq:minimax_obj} for DAF, RDA-ADDA;
    \item In RDA-MMD, the factor of the MMD item in the objective selected from $\{0.1, 1.0, 10.0\}$.
\end{itemize}
For RDA-DANN, we set a schedule $$\lambda = \frac{2}{1+\exp(-10e/E)}-1,$$ where $e$ denotes the current epoch and $E$ denotes the total number of epochs for the factor $\lambda$ of the reversed gradient from the domain discriminator according to \citet{ganin_domain-adversarial_2016}.

\Tabref{tab:DAF_hps} summarizes the specific configurations of the hyper-parameters for our proposed DAF model in the experiments.
\begin{table}[h]
\centering
\begin{tabular}{|c|c|c|c|c|c|}
\hline
                & $h$                 & $l_{\text{MLP}}$   & $s$                    & $\gamma$              & $\lambda$            \\ \hline
cold-start      & \multirow{2}{*}{64} & \multirow{2}{*}{1} & \multirow{2}{*}{(3,5)} & \multirow{2}{*}{1e-3} & \multirow{2}{*}{1.0} \\ \cline{1-1}
few-shot        &                     &                    &                        &                       &                      \\ \hline
\textit{elec}  & 128                 & 1                  & 13                     & 1e-3                  & 1.0                  \\ \hline
\textit{traf}& 64                  & 1                  & (3,17)                 & 1e-2                  & 10.0                 \\ \hline
\textit{wiki}  & 64                  & 2                  & (3,5)                  & 1e-3                  & 1.0                  \\ \hline
\textit{sales} & 128                 & 2                  & (3,5)                  & 1e-3                  & 1.0                  \\ \hline
\end{tabular}
\caption{Hyperparameters of DAF models in various synthetic and real-world experiments.}
\label{tab:DAF_hps}
\end{table}

The models are trained for at most $100K$ iterations, which we empirically find to be more than sufficient for the models to converge. We use early stopping with respect to the ND metric on the validation set.

\section{Detailed Experiment Results}\label{sec:dr}
\begin{table*}[h]
\centering
\scalebox{0.9}{
    \begin{tabular}{c|cccccc}
    \toprule
    Task    & \multicolumn{3}{c|}{Cold-start}                                                                     & \multicolumn{3}{c}{Few-shot}                                                                        \\ \hline
    $N$     & \multicolumn{3}{c|}{5000}                                                                           & 20                                            & 50                       & 100                      \\ \hline
    $T$     & 36                       & 45                       & \multicolumn{1}{c|}{54}                       & \multicolumn{3}{c}{144}                                                                             \\ \hline
    $\tau$  & \multicolumn{6}{c}{18}                                                                                                                                                                                    \\ \midrule
    DeepAR  & 0.053$\pm$0.003          & 0.037$\pm$0.002          & \multicolumn{1}{c|}{\textbf{0.031$\pm$0.002}} & 0.062$\pm$0.003                               & 0.059$\pm$0.004          & 0.059$\pm$0.003          \\
    N-BEATS & 0.044$\pm$0.001          & 0.044$\pm$0.001          & \multicolumn{1}{c|}{0.042$\pm$0.001}          & 0.079$\pm$0.001                               & 0.060$\pm$0.001          & 0.054$\pm$0.002          \\
    CT      & 0.042$\pm$0.001          & 0.041$\pm$0.004          & \multicolumn{1}{c|}{0.038$\pm$0.005}          & 0.095$\pm$0.003                               & 0.074$\pm$0.005          & 0.071$\pm$0.002          \\
    AttF    & 0.042$\pm$0.001          & 0.041$\pm$0.004          & \multicolumn{1}{c|}{0.038$\pm$0.005}          & 0.095$\pm$0.003                               & 0.074$\pm$0.005          & 0.071$\pm$0.002          \\ \midrule
    MetaF   & 0.045$\pm$0.005          & 0.043$\pm$0.006          & \multicolumn{1}{c|}{0.042$\pm$0.002}          & 0.071$\pm$0.004                               & 0.061$\pm$0.003          & 0.053$\pm$0.003          \\
    PTF     & 0.039$\pm$0.006          & 0.037$\pm$0.005          & \multicolumn{1}{c|}{0.034$\pm$0.008}          & 0.086$\pm$0.004                               & 0.086$\pm$0.003          & 0.081$\pm$0.005          \\
    DATSING & 0.039$\pm$0.004          & 0.039$\pm$0.002          & \multicolumn{1}{c|}{0.037$\pm$0.001}          & 0.078$\pm$0.005                               & 0.076$\pm$0.006          & 0.058$\pm$0.005          \\
    RDA-ADDA     & \textbf{0.035$\pm$0.002} & 0.034$\pm$0.001          & \multicolumn{1}{c|}{0.034$\pm$0.001}          & \textbf{0.059$\pm$0.003}                      & \textbf{0.054$\pm$0.003} & 0.053$\pm$0.007          \\
    DAF     & \textbf{0.035$\pm$0.003} & \textbf{0.030$\pm$0.003} & \multicolumn{1}{c|}{\textbf{0.029$\pm$0.003}} & \multicolumn{1}{c|}{\textbf{0.057$\pm$0.004}} & \textbf{0.055$\pm$0.001} & \textbf{0.051$\pm$0.001} \\ \bottomrule
    \end{tabular}}
\caption{Performance comparison of DAF to all baselines on synthetic datasets. The winners and the competitive followers (the gap is smaller than its standard deviation over 5 runs) are bolded for reference.}
\label{tab:full_syn_result}
\end{table*}

\begin{table*}[h]
\centering
\scalebox{0.7}{
    \begin{tabular}{c|cc|cc|cc|cc}
    \toprule
    $\gD_{\gT}$    & \multicolumn{2}{c|}{\textit{traf}}                                                & \multicolumn{2}{c|}{\textit{elec}}                            & \multicolumn{2}{c|}{\textit{wiki}}                            & \multicolumn{2}{c}{\textit{sales}}                            \\ \hline
    $\gD_{\gS}$    & \multicolumn{1}{c|}{\textit{elec}}                     & \textit{wiki}                     & \multicolumn{1}{c|}{\textit{traf}} & \textit{sales}                    & \multicolumn{1}{c|}{\textit{traf}} & \textit{sales}                    & \multicolumn{1}{c|}{\textit{elec}} & \textit{wiki}                     \\ \midrule
    DAR      & \multicolumn{2}{c|}{0.205$\pm$0.015}                                     & \multicolumn{2}{c|}{0.141$\pm$0.023}                 & \multicolumn{2}{c|}{0.055$\pm$0.010}                 & \multicolumn{2}{c}{0.305$\pm$0.005}                  \\
    N-BEATS   & \multicolumn{2}{c|}{0.191$\pm$0.003}                                     & \multicolumn{2}{c|}{0.147$\pm$0.004}                 & \multicolumn{2}{c|}{0.059$\pm$0.008}                 & \multicolumn{2}{c}{0.299$\pm$0.005}                  \\
    VT       & \multicolumn{2}{c|}{0.187$\pm$0.003}                                     & \multicolumn{2}{c|}{0.144$\pm$0.004}                 & \multicolumn{2}{c|}{0.061$\pm$0.008}                 & \multicolumn{2}{c}{0.293$\pm$0.005}                  \\
    CT       & \multicolumn{2}{c|}{0.183$\pm$0.013}                                     & \multicolumn{2}{c|}{0.131$\pm$0.005}                 & \multicolumn{2}{c|}{0.051$\pm$0.006}                 & \multicolumn{2}{c}{0.324$\pm$0.013}                  \\
    AttF     & \multicolumn{2}{c|}{0.182$\pm$0.007}                                     & \multicolumn{2}{c|}{0.137$\pm$0.005}                 & \multicolumn{2}{c|}{0.050$\pm$0.003}                 & \multicolumn{2}{c}{0.308$\pm$0.002}                  \\ \midrule
    MetaF    & \multicolumn{1}{c|}{0.190$\pm$0.005}          & 0.188$\pm$0.002          & 0.151$\pm$0.004           & 0.144$\pm$0.004          & 0.061$\pm$0.003           & 0.059$\pm$0.005          & 0.311$\pm$0.001           & 0.329$\pm$0.002          \\
    PTF      & \multicolumn{1}{c|}{0.184$\pm$0.003}          & 0.185$\pm$0.004          & 0.144$\pm$0.005           & 0.138$\pm$0.007          & \textbf{0.044$\pm$0.003}  & \textbf{0.047$\pm$0.002} & 0.287$\pm$0.004           & 0.292$\pm$0.007          \\
    DATSING  & \multicolumn{1}{c|}{0.184$\pm$0.004}          & 0.189$\pm$0.005          & 0.137$\pm$0.003           & 0.149$\pm$0.009          & 0.049$\pm$0.002           & 0.052$\pm$0.004          & 0.301$\pm$0.008           & 0.305$\pm$0.008          \\
    RDA-DANN & \multicolumn{1}{c|}{0.181$\pm$0.009}          & 0.180$\pm$0.004          & 0.133$\pm$0.005           & 0.135$\pm$0.007          & 0.047$\pm$0.005           & 0.053$\pm$0.002          & 0.297$\pm$0.004           & 0.287$\pm$0.009          \\
    RDA-ADDA & \multicolumn{1}{c|}{0.174$\pm$0.005}          & 0.181$\pm$0.003          & 0.134$\pm$0.002           & 0.142$\pm$0.003          & \textbf{0.045$\pm$0.003}  & \textbf{0.049$\pm$0.003} & \textbf{0.281$\pm$0.001}  & \textbf{0.287$\pm$0.002} \\
    RDA-MMD  & \multicolumn{1}{c|}{0.186$\pm$0.004}          & 0.179$\pm$0.004          & 0.140$\pm$0.006           & 0.144$\pm$0.003          & \textbf{0.045$\pm$0.003}  & 0.052$\pm$0.004           & 0.291$\pm$0.004           & 0.289$\pm$0.003          \\
    DAF      & \multicolumn{1}{c|}{\textbf{0.169$\pm$0.002}} & \textbf{0.176$\pm$0.004} & \textbf{0.125$\pm$0.008}  & \textbf{0.123$\pm$0.005} & \textbf{0.042$\pm$0.004}  & \textbf{0.049$\pm$0.003} & \textbf{0.277$\pm$0.005}  & \textbf{0.280$\pm$0.007} \\ \bottomrule
    \end{tabular}
}
\caption{Performance comparison of DAF to all baselines on real-world benchmark datasets. The winners and the competitive followers (the gap is smaller than its standard deviation over 5 runs) are bolded for reference.}
\label{tab:full_real_result}
\end{table*}

Tables \ref{tab:full_syn_result}-\ref{tab:full_real_result} display a comprehensive comparison of DAF with all the aforementioned baselines on the synthetic and real-world data, respectively. As a conclusion, we see that DAF outperforms or is on par with the baselines in all cases with RDA-ADDA being the most competitive in some cases.



\begin{figure*}[b]
    \centering
    \includegraphics[scale=0.3]{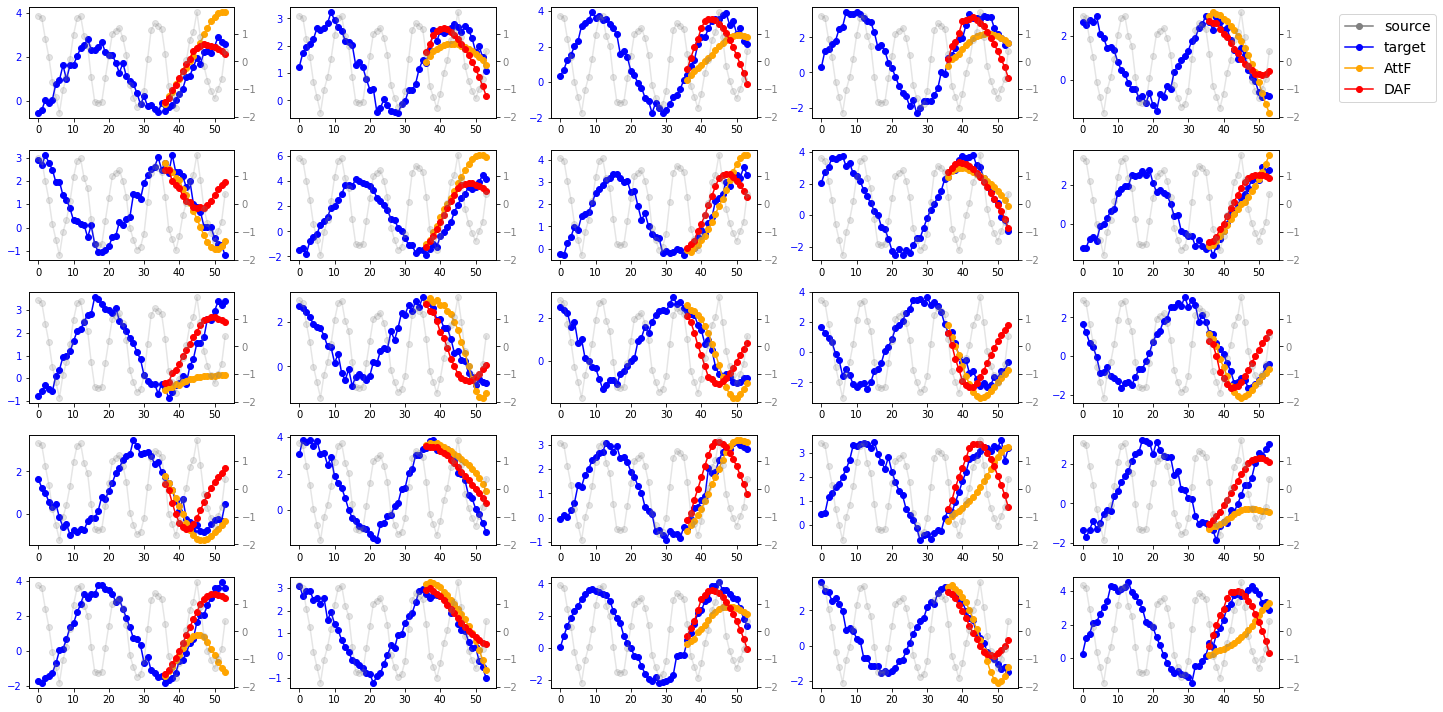}
    \caption{Test samples in the synthetic few-shot experiment where $N=20$.   The y-axis corresponding to the source is shown in grey, and that for the target in blue.}
    \label{fig:syn_samples}
\end{figure*}

\begin{figure*}[b]
    \centering
    \includegraphics[scale=0.3]{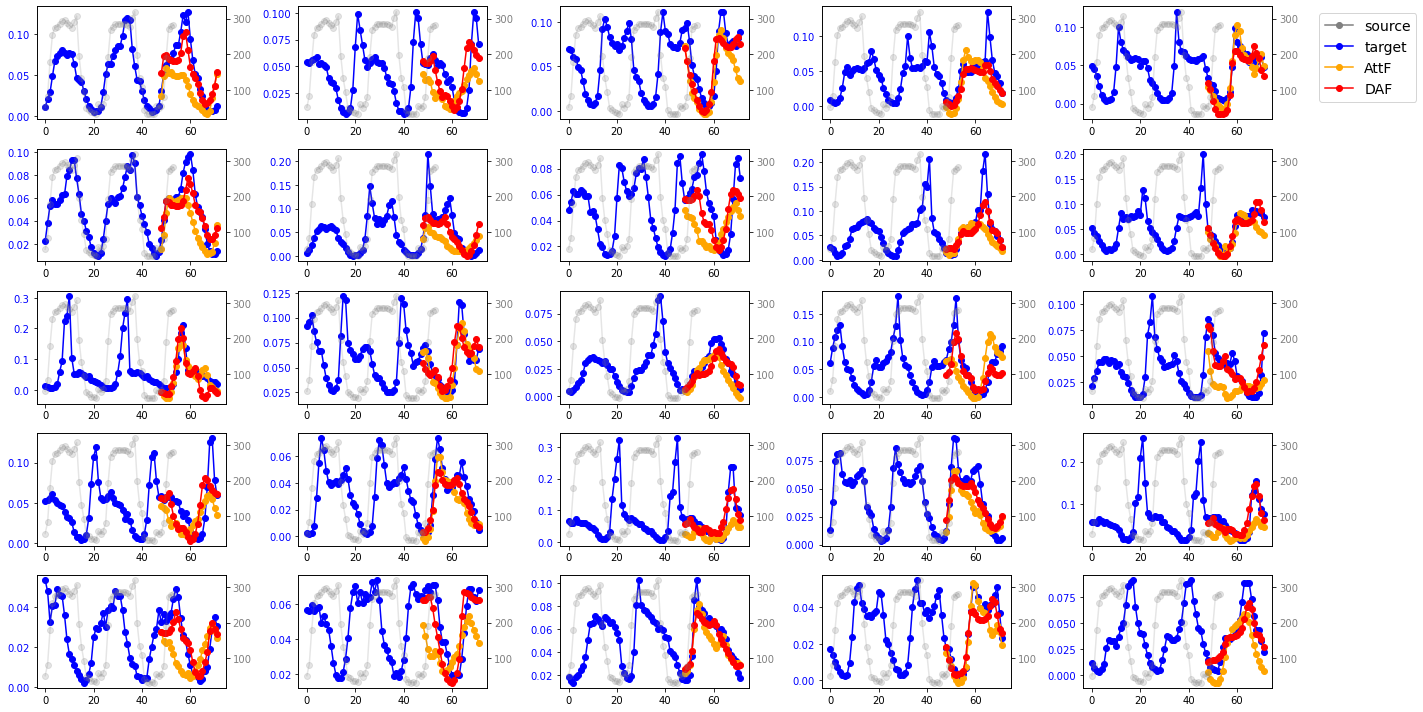}
    \caption{Test samples with source data \textit{elec}  and target data \textit {traf} experiment. The y-axis corresponding to the source is shown in grey, and that for the target in blue.}
    \label{fig:e2t_samples}
\end{figure*}

We also provide visualizations of forecasts for both synthetic and real-world experiments. \Figref{fig:syn_samples} provides more samples for the few-shot experiment where $N=20$ as a complement to Figure 4. 
We see that DAF is able to approximately capture the sinusoidal signals even if the input is contaminated by white noise in most cases, while AttF fails in many cases. \Figref{fig:e2t_samples} illustrates the performance gap between DAF and AttF in the experiment with source data \textit{elec} and target data \textit{traf} as an example in real scenarios. While AttF generally captures daily patterns, DAF performs significantly better.

